*coatings*

*Article*

# Ensemble of Deep Convolutional Neural Networks for Automatic Pavement Crack Detection and Measurement

**Zhun Fan** [1,2], **Chong Li** [1,2,3,*], **Ying Chen** [1,2], **Paola Di Mascio** [3], **Xiaopeng Chen** [4], **Guijie Zhu** [1,2] and **Giuseppe Loprencipe** [3,*]

1   Key Lab of Digital Signal and Image Processing of Guangdong Province, Shan'tou 515063, China; zfan@stu.edu.com (Z.F.); 19ychen1@stu.edu.cn (Y.C.); 16gjzhu@stu.edu.cn (G.Z.)
2   College of Engineering, Shantou University, Shan'tou 515063, China
3   Department of Civil, Construction and Environmental Engineering, Sapienza University of Rome, Rome 00184, Italy; paola.dimascio@uniroma1.it (P.D.M.); giuseppe.loprencipe@uniroma1.it (G.L.)
4   Department of Industrial Engineering, Pusan National University, Busan 609735, Korea; xiaopengchen388@gmail.com
*   Correspondence: 15cli@stu.edu.cn (C.L.); giuseppe.loprencipe@uniroma1.it (G.L.)

**Abstract:** Automated pavement crack detection and measurement are important road issues. Agencies have to guarantee the improvement of road safety. Conventional crack detection and measurement algorithms can be extremely time-consuming and low efficiency. Therefore, recently, innovative algorithms have received increased attention from researchers. In this paper, we propose an ensemble of convolutional neural networks (without a pooling layer) based on probability fusion for automated pavement crack detection and measurement. Specifically, an ensemble of convolutional neural networks was employed to identify the structure of small cracks with raw images. Secondly, outputs of the individual convolutional neural network model for the ensemble were averaged to produce the final crack probability value of each pixel, which can obtain a predicted probability map. Finally, the predicted morphological features of the cracks were measured by using the skeleton extraction algorithm. To validate the proposed method, some experiments were performed on two public crack databases (CFD and AigleRN) and the results of the different state-of-the-art methods were compared. To evaluate the efficiency of crack detection methods, three parameters were considered: precision ($Pr$), recall ($Re$) and F1 score ($F1$). For the two public databases of pavement images, the proposed method obtained the highest values of the three evaluation parameters: for the CFD database, $Pr = 0.9552$, $Re = 0.9521$ and $F1 = 0.9533$ (which reach values up to 0.5175 higher than the values obtained on the same database with the other methods), for the AigleRN database, $Pr = 0.9302$, $Re = 0.9166$ and $F1 = 0.9238$ (which reach values up to 0.7313 higher than the values obtained on the same database with the other methods). The experimental results show that the proposed method outperforms the other methods. For crack measurement, the crack length and width can be measure based on different crack types (complex, common, thin, and intersecting cracks.). The results show that the proposed algorithm can be effectively applied for crack measurement.

## 1. Introduction

A pavement crack is a critical failure of a pavement structure and presents a potential threat to road and highway safety [1–6]. Road crack detection and measurement play a role in road management [7–9]. Conventional crack detection and measurement algorithms are time-consuming and less efficient. Therefore, automated crack detection and measurement outperform the conventional methods and for this reason, they have received increased attention from researchers.

In a structure, the geometry of the cracks (width, length, and orientation) can be retrieved from images and the results can be used to evaluate the required safety and maintenance work. This process is similar to the automated road crack detection one and similar method to detect cracks [10] can be used. Oh et al. proposed a bridge detection system based on a robot, including a specially designed car, a robot mechanism, a control system for mobility, and a machine vision system [11].

In recent decades, some algorithms for image processing have been widely used to detect road cracks. In early studies, many researchers adopted methods related to threshold [12], edge detection [13,14] and morphology [15] based on photometric and geometric hypotheses to improve the accuracy of road crack detection.

The cracks and background are segmented by using a threshold value [12–16]. Some researchers applied Canny and Sobel methods to detect cracks [13,14]. In other studies, the geometric information of the cracks was taken into consideration to reduce false detection [15]. The wavelet transform method was employed to detect crack regions by Subirats et al. [17]. Although these methods can be used to detect cracks, they cannot find all the cracks as a consequence of noise interference. Recently, alternative analytical methods have been presented to improve the performance of crack detection:

- Minimal Path Methods: the principle of this approach is to suppose that minimal paths in the image correspond to road cracks. Kaul et al., in [18], proposed a new algorithm to detect crack curve with unknown endpoints and topology based on minimal path. Nguyen et al., in [19], applied the Free-From Anisotropy to address brightness and connectivity issues in the cracks. Amhaz et al., in [20], considered the local and global level to choose endpoints and minimal path for crack detection, using two-dimensional pavement images.

- Machine Learning: Recently, many algorithms have been proposed to detect cracks based on machine learning. A support vector machine (SVM) was employed to detect aircraft skin cracks [21]. Oliveira and Correia, in [22], proposed an unsupervised learning algorithm named CrackIT to detect cracks. After that, they developed research to extend their work to the CrackIT toolbox [23]. A new descriptor with a random structure forests algorithm has been proposed to describe crack and non-crack pixels [24]. Due to overlay depending on feature descriptors and complex road conditions, it is difficult for operators to inspect road cracks.

- Deep Learning: For multi-class classification tasks, deep learning has presented a better performance than traditional algorithms. Moreover, many distress detection issues adopted the deep learning to inspect and recognize cracks. Cha et al. used the convolutional neural networks (CNN) and Faster-RCNN to detect road cracks [25,26]. In CrackNet [27], the proposed CNN without pooling layers was used to inspect cracks and improve accuracy. In CrackNet-R [28], Zhang et al. proposed a Gated Recurrent Multilayer Perception (GRMLP), which was embedded into the CNN to perform automated pavement crack detection. A structured prediction method with CNN was proposed to inspect cracks pixels [29]. Yang et al. in [30] adopted the Fully Convolutional Network (FCN) to perform automated road crack detection and measurement.

- Ensemble Learning: An ensemble network was proposed to perform medical image classification [31]. Wen et al. designed an ensemble network based on probability fusion for facial expression recognition [32]. Maji et al. proposed an ensemble network to detect retinal vessels [33].

The width of the predicted crack image can be measured by extracting the morphological aspects, which can be segmented into a thinned crack skeleton. Many algorithms can be employed to skeletonize the predicted images, including the 3D-medial axis thinning algorithm, medial-axis and Hilditch's algorithm [34,35]. A crack defragmentation technique was proposed to measure the average crack width [36].

Inspired by the above observations, we propose an ensemble network (without a pooling layer) based on the probability fusion for automated pavement crack detection and measurement, shown in Figure 1. Specifically, an ensemble network was employed to evaluate the small cracks' structure from raw images, shown in Figure 1a. The individual CNN model adopts a structured prediction method to detect cracks. The outputs of the individual convolutional neural network model for the ensemble (Figure1b) were averaged to produce the final crack probability of each pixel, which can obtain a predicted probability map, shown in Figure 1c. The segmentation image was obtained after the morphological operation shown in Figure 1d. Then, the crack skeleton can be obtained based on the medial-axis algorithm, shown in Figure 1e. Finally, the predicted morphological features of cracks were measured by using the skeleton extraction algorithm, shown in Figure 1f.

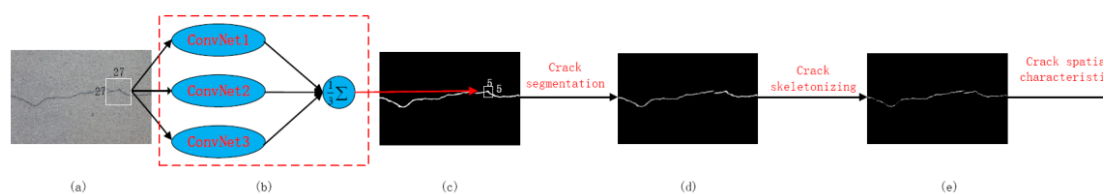

**Figure 1.** Overview of the automated pavement crack detection and measurement system, (a) The raw image; (b) The ensemble network model; (c) The output image of ensemble network; (d) The crack segmentation image; (e) The extracted crack skeleton based on the medial-axis algorithm; (f) The extracted crack width and length.

The contributions of an ensemble network are listed below:

1. We propose an ensemble network based on probability fusion for automated pavement crack detection and measurement.
2. The structured predicted method was embedded into individual CNNs for an ensemble network. The designed individual CNNs can improve the accuracy of crack detection by discarding the pooling layers.
3. The designed ensemble neural network model was employed to obtain a satisfactory accuracy for crack detection.
4. The crack width and length can be measured based on the predicted crack maps.

The rest of this paper is organized as follows. The details of the proposed ensemble of convolution neural networks are described in Section II. Then, we conducted comprehensive experiments to show the crack detection and measurement performance for the proposed method and compared them with other algorithms in Section III. Finally, the conclusions are provided in Section IV.

## 2. Methods

This section introduces the details of the ensemble network for automated pavement crack detection and measurement.

### 2.1. Convolutional Neural Networks

The CNN shows that the network employs a mathematical operation, named convolution, which is a specialized type of linear operation. Convolutional networks are simply neural networks that use convolution in place of general matrix multiplication in at least one of their layers [37]. The CNN have four different parts, including a convolution, pooling, full connection and activation function.

Convolution layer: It contains a K filter (or kernels) with the weight W. By applying the following equation, the convolution process can be adopted to obtain the output K of the feature maps:

$$H_i^l = H^{l-1} \bigotimes W_i^l + b_i^l \tag{1}$$

where $b_i^l$ and $W_i^l$ are the bias and weights of the $i^{th}$ filter based on the $l^{th}$ convolutional layer, respectively. $H_i^l$ is the feature maps.

Full connection layer: $b^l$ and $W^l$ are the bias and weights for the full connection layer $l^{th}$. The operating process for the full connection layer is shown in the following equation:

$$H^l = flatten(H^{l-1}) * W^l \oplus b^l \tag{2}$$

where $flatten(\cdot)$ indicates that the feature maps are tiled along the height value. The symbols * and $\oplus$ represent, respectively, the matrix multiplication and element-wise addition.

Pooling layer: Pooling was employed to decrease the size of image, which contains two types: max pooling and average pooling. When a sliding window moves and scans the feature maps, the average value can be obtained for average pooling. Therefore, the max pooling can calculate the maximum value.

Activation function: The activation function rectified linear unit (ReLU) [38] was employed to increase the non-linearity of the output after the convolution process. The activation functions *sigmoid* and *softmax* are usually applied to the binary classification and multi-label classification [39], respectively.

### 2.2. Structured Prediction Method

The structured prediction method was applied into the individual network of the ensemble network, which was proposed by our original article [29].

The architecture of an individual network of ensemble network is shown in Figure 2 [29]. The size of the input patch is 27 × 27 with three channels, and the size of 5 × 5 is defined as out structure, which is shown in Figure 2.

The feature maps are shown with the cubes, which were obtained after the convolution operation. The circles indicate the full connection layers that were used to obtain the output. The layer names are followed by numbers of feature maps, specifying kernel size, stride and padding (see Figure 2).

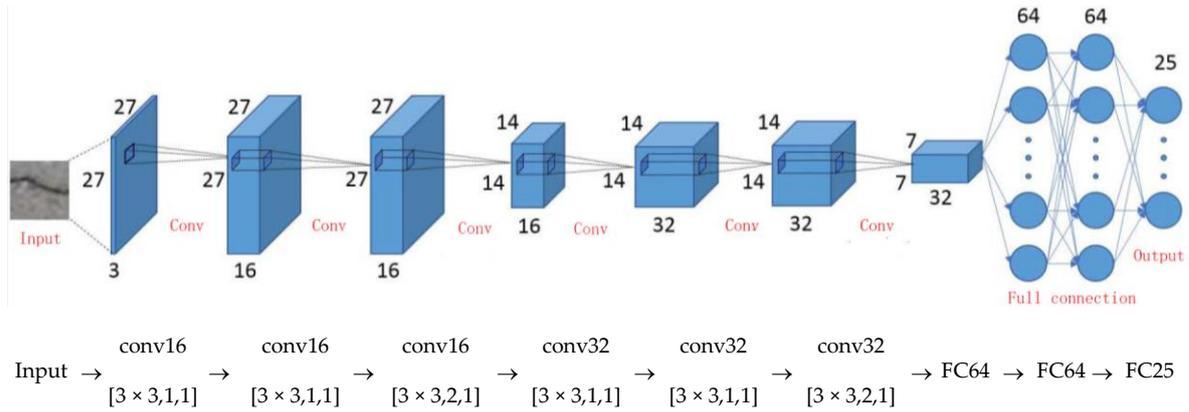

**Figure 2.** The architecture of an individual convolutional neural network (CNN).

In this architecture of the CNN, the kernel of 3 × 3 was employed to perform a convolution operation which was verified in VGG-net [40]. The zeros padding operation was employed to obtain the spatial resolution of the feature maps during the convolution process. In order to increase the level of abstraction for the feature maps, the input images were downsized by using the pooling layer, which leads to the loss of input information [27,41]. Therefore, we discarded the pooling layer in this CNN model.

The output patch 5 × 5 is a structured prediction center based on the input patch 27 × 27, shown in Figure 2. The 25 neurons were obtained from the 5 × 5 windows, which is the output layer of the CNN model. The output size of the corresponding ground truth is 5 × 5 labels. Pavement crack detection is a binary classification task. Therefore, the activation function *sigmoid* was used for the final output for the binary classification task. The ReLU was employed to increase the non-linearity for hidden layers.

In the CNN training process, we adopted the cross entropy loss function to minimize classification error with the following the equation:

$$L = -\sum_{i=1}^{s^2}(y_i \, log \, \hat{y}_i + (1 - y_i) \, log(1 - \hat{y}_i)) \tag{3}$$

where $y_i$ and $\hat{y}_i$ are the label for ground truth and the prediction value based on the $i^{th}$ output, respectively. The number of labels is defined as $s^2$. At the same time, the weight decay is employed to penalize weight factors of the CNN model to avoid the network overfitting. Therefore, we applied the $L_2$ penalty term into the loss function. The total loss $L_{total}$ is shown with the following equation:

$$L_{total} = L + \beta \cdot \frac{1}{2}\sum_{j} W_j^2 \tag{4}$$

where $\beta$ is the penalty factor for $L_2$ and $W_j$ is the $j^{th}$ weight for CNN. Then, we also adopt the Dropout to prevent network overfitting [42].

## 2.3. Ensemble Network Learning Method

Ensemble network learning is an algorithm that constructs multiple models to address the artificial intelligence task [43]. The ensemble learning method is able to promote performance among the CNN models and decrease network overfitting, which can combine various classifiers to achieve a better performance than a single classifier.

The outputs of individual CNN model based on ensemble network are averaged to produce the final prediction of crack probability. Specifically, if the ensemble network contains $k$ models $m_1, ..., m_k$, the output probability $p(x = y_i | m_j)$ presents that the output data $x$ is classified as $y_i$ based on model $m_j$, and the final ensemble prediction is shown with the following equation:

$$p(x = y_i | m_1, \ldots, m_k) = \frac{1}{k} \sum_{j=1}^{k} p(x = y_i | m_j) \qquad (5)$$

The ensemble network is able to show a better performance and higher accuracy for crack detection than the individual models.

### 2.4. Crack Measurement

Once we obtained the detected crack images by the ensemble network, the morphological aspects could be calculated and extracted from binary crack images. As shown in Figure 1, the predicted binary crack images are labeled to generate the segmented crack images. In order to generate crack skeletons images, we thinned the segmented crack images and used one pixel to show the crack skeleton. Finally, we can obtain crack morphological features based on crack skeleton images [30].

### 2.4.1. Crack Segmentation

In order to separate crack pixels from the background, we need to label each crack pixels from the crack images to segment them. This operation contains three steps: filling small holes, discarding noisy pixels, and labeling each crack pixel.

The closing operating based on morphological operation is employed to fill small holes in a crack image, formulated with the following equation:

$$closing: \big( (f \oplus \psi) \ominus \psi \big) \qquad (6)$$

where $\oplus$ and $\ominus$ are the dilation and erosion operations for the morphological operation, respectively. $f$ and $\psi$ are the crack image and the structure element, respectively. The operation $\oplus$ is used to increase the regions of the crack pixels and the operation $\ominus$ erodes the boundary regions of the crack pixels.

The opening operation based on the morphological operation is applied to warp off noisy pixels. Compared with the closing operation, the opening operation has an adverse order with regards to the erosion and dilation operation. The opening operation is defined as

$$opening: \big( (f \ominus \psi) \oplus \psi \big) \qquad (7)$$

Image segmentation is the attribution of different labels to different regions of an image. Therefore, we labelled the individual cracks to generate segmentation images. The crack images can be segmented after labeling individual cracks.

### 2.4.2. Crack Skeleton

The goal of the crack skeletons is to use the crack of a single pixel to visualize the cracks' topology. The extracted crack skeletons can be used as a reference value for the structural health monitoring and road maintenance. In this article, we employed a medial-axis algorithm to perform crack skeletonization, which can realize real-time detection [30,35].

When we obtain the crack skeletons with single pixel wide, the following equation can be used to calculate the length of the cracks:

$$L_{crack} = \sum f(x, y) dl \qquad (8)$$

where *f(x, y)* and *dl* are the calibrated displacements of the pixels in the crack images and the finite length of the crack skeleton elements, respectively. In this project, we assume that there is no geometric distortion. Hence, *f(x, y)* is defined as unique. At the same time, we can calculate length of cracks by the displacement of pixels for crack skeletons. The average width of cracks can be ormulated with the following equation:

$$W_{avg} = \frac{\sum f^2(x,y)ds}{L_{crack}}$$ (9)

where *ds* is the finite area of the crack elements. Therefore, we can obtain the physical length and width of a crack according to the image resolution. These reference values can help engineers to evaluate and estimate the security performance of a pavement.

## 3. Experimental Results

In this section, we mainly discuss implementation details for an ensemble network and present the experimental results.

### 3.1. Training and Testing

An ensemble of convolutional neural networks was programmed by Tensorflow based on the Linux system. The experimental results are implemented on the GPU workstation equipped with the types of NVIDIA-GTX-Titan-XP.

In this project, the public databases CFD [24] and AigleRN [44] were employed to evaluate an ensemble network. The CFD includes 118 images with pixel size of 320 x 480, and it was obtained using an iPhone 5 smartphone ( Apple Inc. State of California, USA) on a pavement in Beijing, China. The CFD database contains various types of noise, such as oil spots and shadows. This database is divided into two parts: training set (72 images) and test set (46 images). The AigleRN database has 38 images with a gray level and has two types of resolution (991 × 462 and 311 × 462). This database has complex structures, which were taken from a French pavement. In this case, 24 images were used for training and 14 images were used for testing.

In this project, we employed three numbers to evaluate the accuracy of the model: precision (*Pr*), recall (*Re*), F1 score (*F1*); these parameters are defined with the following equations:

$$Pr = \frac{TP}{TP + FP}$$ (10)

$$Re = \frac{TP}{TP + FN}$$ (11)

$$F1 = \frac{2 \times Pr \times Re}{Pr + Re}$$ (12)

where *TP* is short for the number of true positives. False positive is defined as *FP*. *FN* presents false negatives. For the ensemble networks, when we calculate the metric for *TP*, *FP*, and *FN*, we consider the transitional areas between non-crack and crack pixels. Therefore, a two-pixels distance between the prediction image and the ground truth is accepted in [20,29,45,46]. In this project, we accepted a two-pixel distance.

### 3.2. Ensemble Network

In this sub-section, we mainly explore the number of ensemble networks and thresholds based on CFD and AigleRN databases. The numbers of the neural network are defined as *n* = 1, 3, 5, 7. The decision probability value is defined as *t* = 0.4, 0.5, 0.6, and 0.7 to eliminate low-probability pixels and obtain binary crack images. As is shown in Figures 3 and 4, the *Pr*, *Re*, and *F1* values are shown based on different neural network models and thresholds for the AigleRN and CFD databases.

From Figure 3, it is clear that the experimental results for threshold t = 0.4 and the numbers of neural networks models *n* = 3, have a better performance than the other values based on the AigleRN database. From Figure 4, it can be observed that the experimental results for threshold *t* = 0.6 and the numbers of neural networks models, *n* = 3, have a better performance than the others values based on the CFD database. Therefore, the number of ensemble networks models was set to three in this project, as shown in Figure 1. At the same time, the thresholds were set to *t* = 0.4, 0.6 for the Ensemble Networks and CFD database.

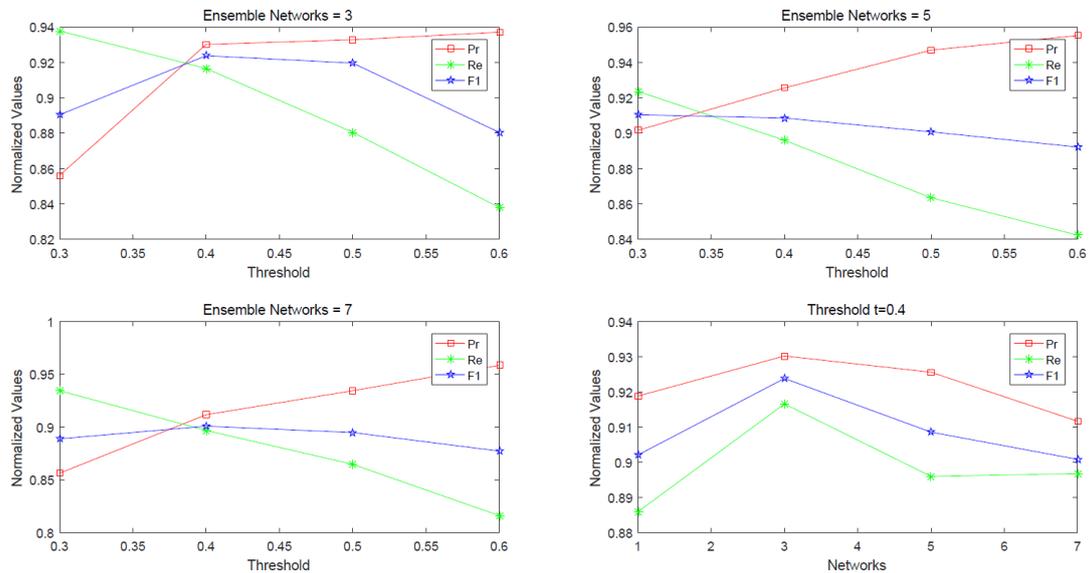

**Figure 3.** The *Pr*, *Re*, and *F1* value variations with different numbers of neural network models and thresholds based on the AigleRN database.

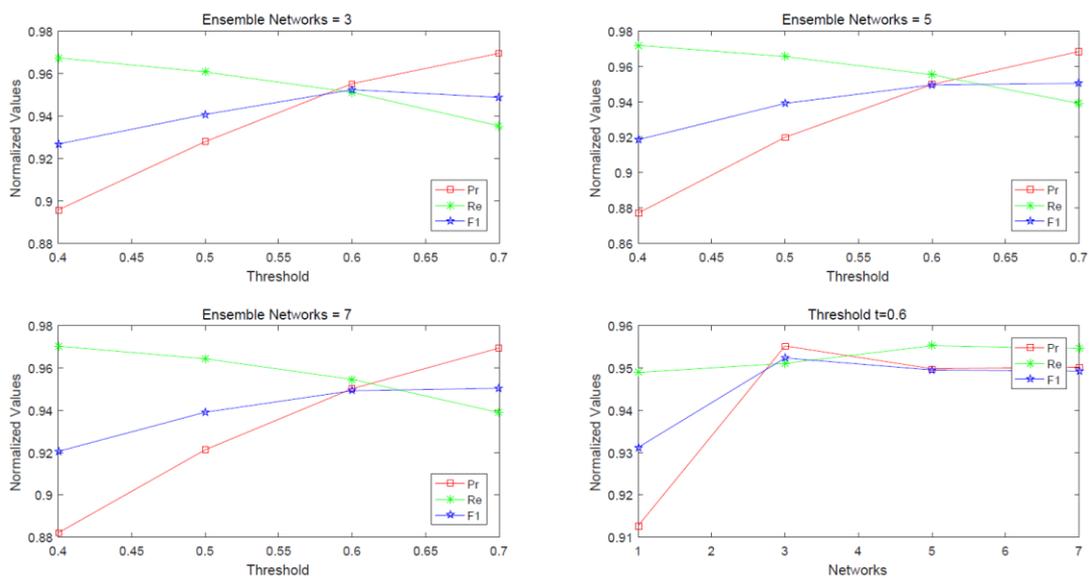

**Figure 4.** The *Pr*, *Re*, and *F1* value variations with different numbers of neural network models and thresholds based on the CFD database.

### 3.3. Results on CFD

Figure 5 and Table 1 present some specimen detections in the public database CFD. The experimental result images for different methods are shown in Figure 5 (from left to right:

original image, ground truth, Canny [13], local threshold [12], CrackForest [24], structured prediction [29], U-net [47], and the proposed method.). It can be observed that noises have a negative influence on the two traditional methods (Canny and local threshold), which cannot be used to detect cracks. It is also clear that CrackForest can obtain wider a crack width than ground truth, with high recall and low precision (recall: 0.9514, precision: 0.7466), as shown in Table 1. This method can overestimate the number of cracks. It is clear that the structured prediction method can obtain several wrong detections, such as white points. This method shows that it is not able to obtain more feature maps for cracks. Although the method of U-net can get the crack skeleton, missed detections also occur in the images.

The experimental results show that ensemble networks can get a more satisfactory accuracy than other algorithms, as shown in Figure 5 and Table 1 (*Pr*: 0.9552, *Re*: 0.9521, *F1*:0.9533). The main reason is that ensemble networks can extract and learn more features than individual networks. Hence, ensemble networks can obtain a satisfactory performance.

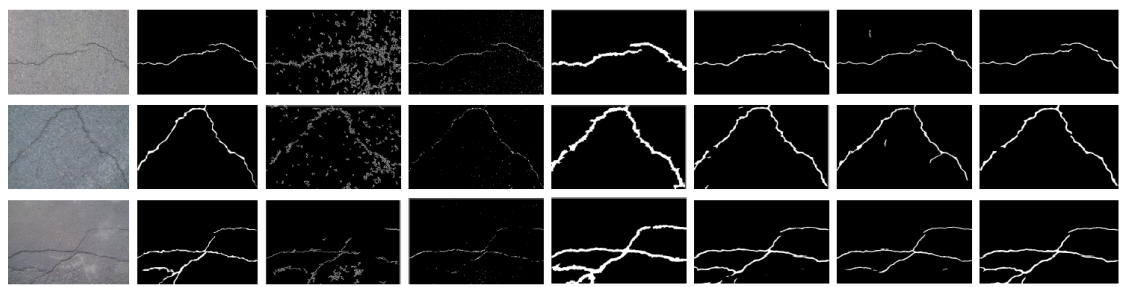

**Figure 5.** Experimental results of the comparison of ensemble networks with other methods based on CFD.

**Table 1.** Crack detection experimental results on CFD.

|  | *Pr* | *Re* | *F1* |
| --- | --- | --- | --- |
| Canny [13] | 0.4377 | 0.7307 | 0.457 |
| Local thresholding [12] | 0.7727 | 0.8274 | 0.7418 |
| CrackForest [24] | 0.7466 | 0.9514 | 0.8318 |
| U-net [47] | 0.9254 | 0.8951 | 0.899 |
| Structured prediction [29] | 0.9119 | 0.9481 | 0.9244 |
| Structured prediction without pooling | 0.9227 | 0.9489 | 0.9312 |
| **Ensemble network** | **0.9552** | **0.9521** | **0.9533** |

### 3.4. Results on AigleRN

Figure 6 and Table 2 show the experimental results with different methods based on AigleRN. The experimental result images for different methods are shown in Figure 6 (from left to right: original image, ground truth, Canny [13], local threshold [12], Free-Form Anisotropy (FFA) [19], Minimal Path Selection (MPS) [18], structured prediction [29], and the proposed method. It is clear that Canny and local threshold are not able to detect continuous cracks and that these methods are sensitive to the noise. The FFA method can inspect some local cracks but also fails to detect continuous cracks. This method is not used to detected global pavement cracks. It can be observed that MPS can find continuous cracks, but the crack skeleton cannot be extracted. The structured prediction methods is effective to inspect the cracks, but there are also missed detections that occur in the images. The ensemble networks can find more continuous cracks and it can extract the crack skeleton, obtaining a good accuracy. Therefore, the proposed ensemble networks method can outperform other algorithms, as shown in Figure 6 and Table 2 (*Pr*:0.9302, *Re*:0.9166, *F1*:0.9238). The main reason is that ensemble networks can extract and learn more features than individual networks. Hence, ensemble networks can obtain a satisfactory performance.

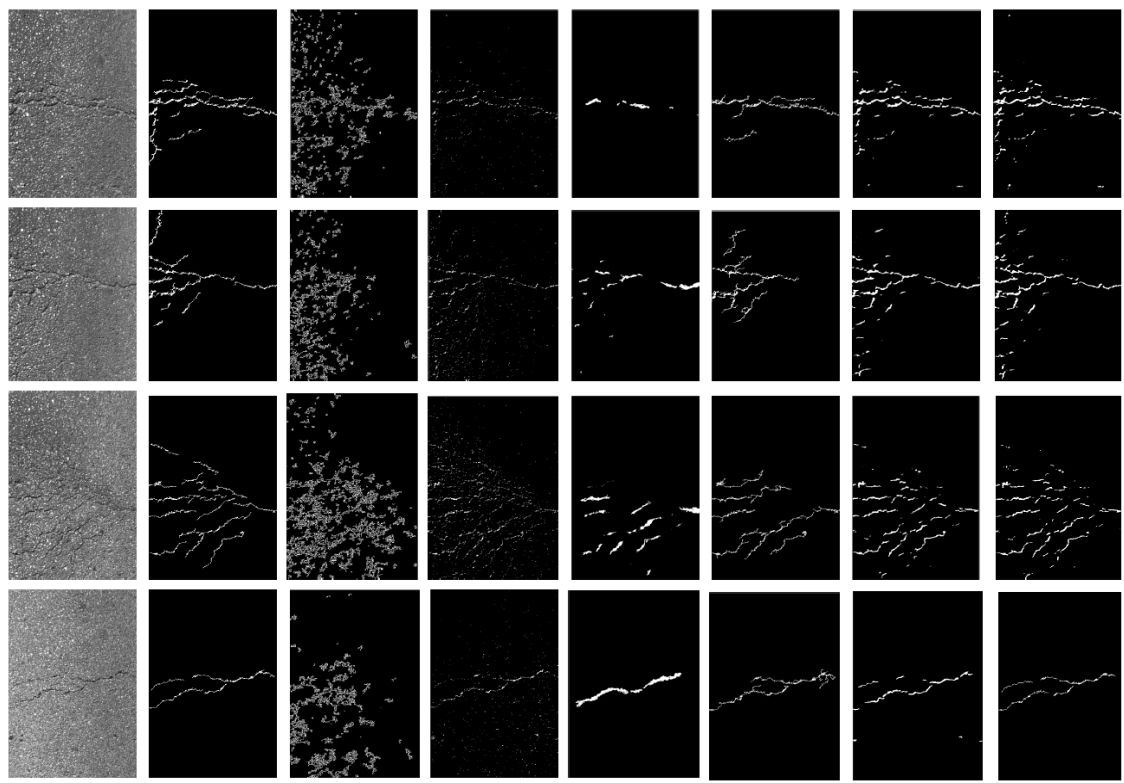

**Figure 6.** Experimental results of the comparison of ensemble networks with other methods based on AigleRN.

**Table 2.** Crack detection experimental results on AigleRN.

|  | Pr | Re | F1 |
| --- | --- | --- | --- |
| Canny [13] | 0.1989 | 0.6753 | 0.2881 |
| Local thresholding [12] | 0.5329 | 0.9345 | 0.667 |
| FFA [19] | 0.7688 | 0.6812 | 0.6817 |
| MPS [18] | 0.8263 | 0.841 | 0.8195 |
| Structured prediction [29] | 0.9178 | 0.8812 | 0.8954 |
| Structured prediction without pooling | 0.9188 | 0.8861 | 0.9021 |
| **Ensemble network** | **0.9302** | **0.9166** | **0.9238** |

### 3.5. Measurements

In this sub-section, we mainly discuss the details of the method implemented for crack measurement and present the main experimental results.

### 3.5.1. Crack Segmentation and Skeleton

As shown in Figure 7, the experimental results show crack segmentation and crack skeleton based on public databases CFD and AigleRN. The images and experimental results from the first row to the third row are based on CFD. The images and experimental results from the fourth row to the sixth row are based on AigleRN. The experimental result images are shown from left to right: original image, ground truth, predicted image, crack segmentation image, and crack skeleton image.

The labels of each cracks are indicated as different colors in the crack segmentation process. The crack skeleton with single-pixel is extracted based on the medial-axis method, which is shown by using different colors. The wider the crack, the lighter the crack skeleton. It is clear that crack segmentation and crack skeleton images are able to present original images based on CFD (from the first row to the third row in Figure 7), which have a better accuracy. The crack skeleton for the complex images based on AigleRN (from the fourth row to the sixth row in Figure 7) can be extracted. Due to the complex structures in the AigleRN database, the output results may have some deficiencies with reference to the public database. However, the disconnected and small cracks can also be detected and skeletonized, which is also shown in the ensemble networks.

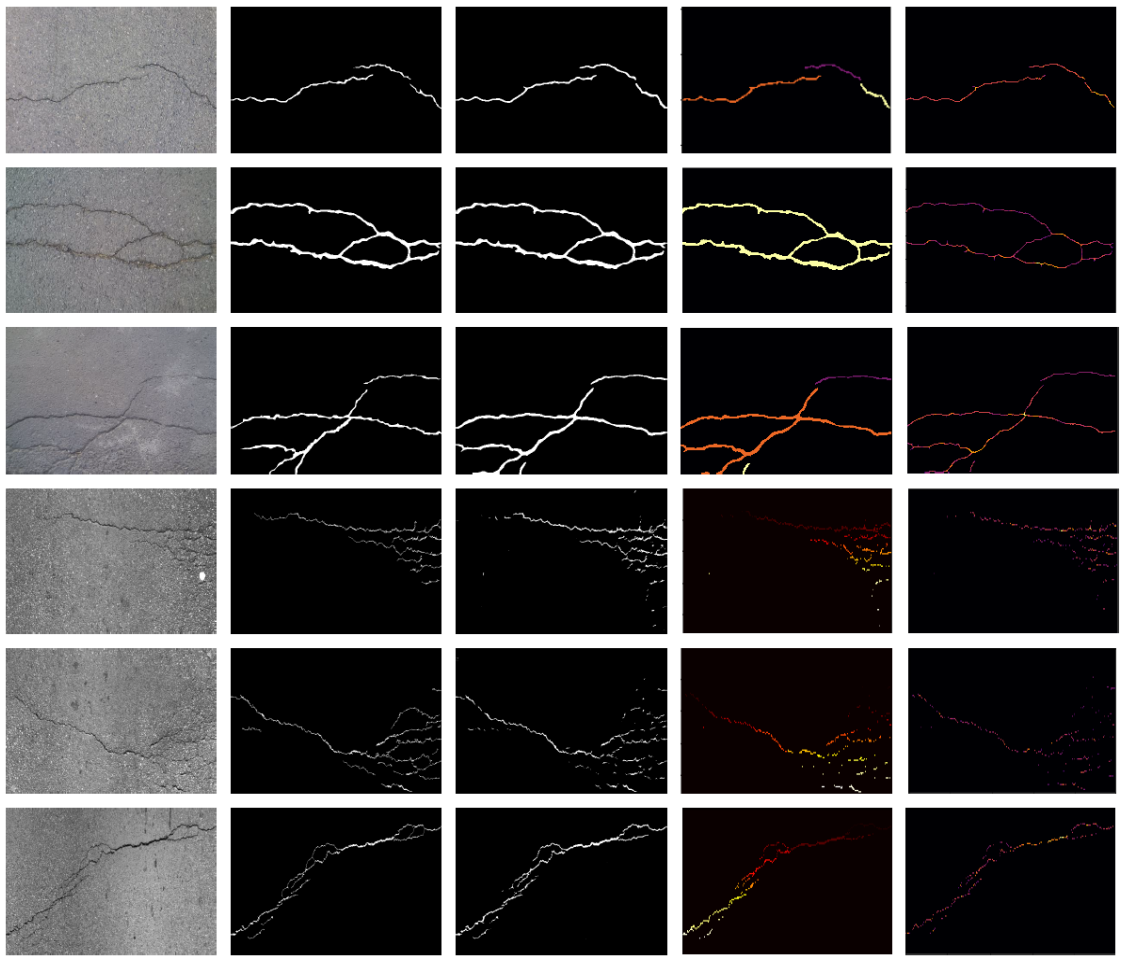

**Figure 7.** The experimental results show crack segmentation and crack skeleton based on public databases CFD and AigleRN.

### 3.5.2. Crack Measurements

Figure 8 presents the morphological features of crack measurement (from left to right: original image, ground truth skeleton, predicted image skeleton, ground truth morphological features, predicted image morphological features.). It is clear that the method tends to overestimate the crack length, shown in Figure 8. The experimental results (Figure 8, first, third, fourth, and sixth rows) show that the ground truth length of the crack skeleton is much larger than the predicted length. The main reason for this is that the disconnected cracks are eliminated and missed for crack detection, which can reduce the numbers of cracks.

The experimental results (Figure 8, second and fifth rows) show that the ground truth length of crack skeleton is much lower than the predicted length, which leads to a larger crack mean width than ground truth. The main reason for this is that we used morphological operations (opening and closing operation) to fill the hole and eliminate the single pixel. These methods can fill the whole crack pixels or the neighbor pixels and eliminate the isolate pixels, which can increase the number of predicted crack pixels. At the same time, the order for the opening and closing methods for the morphological operation may have had an influence on the numbers of crack pixels.

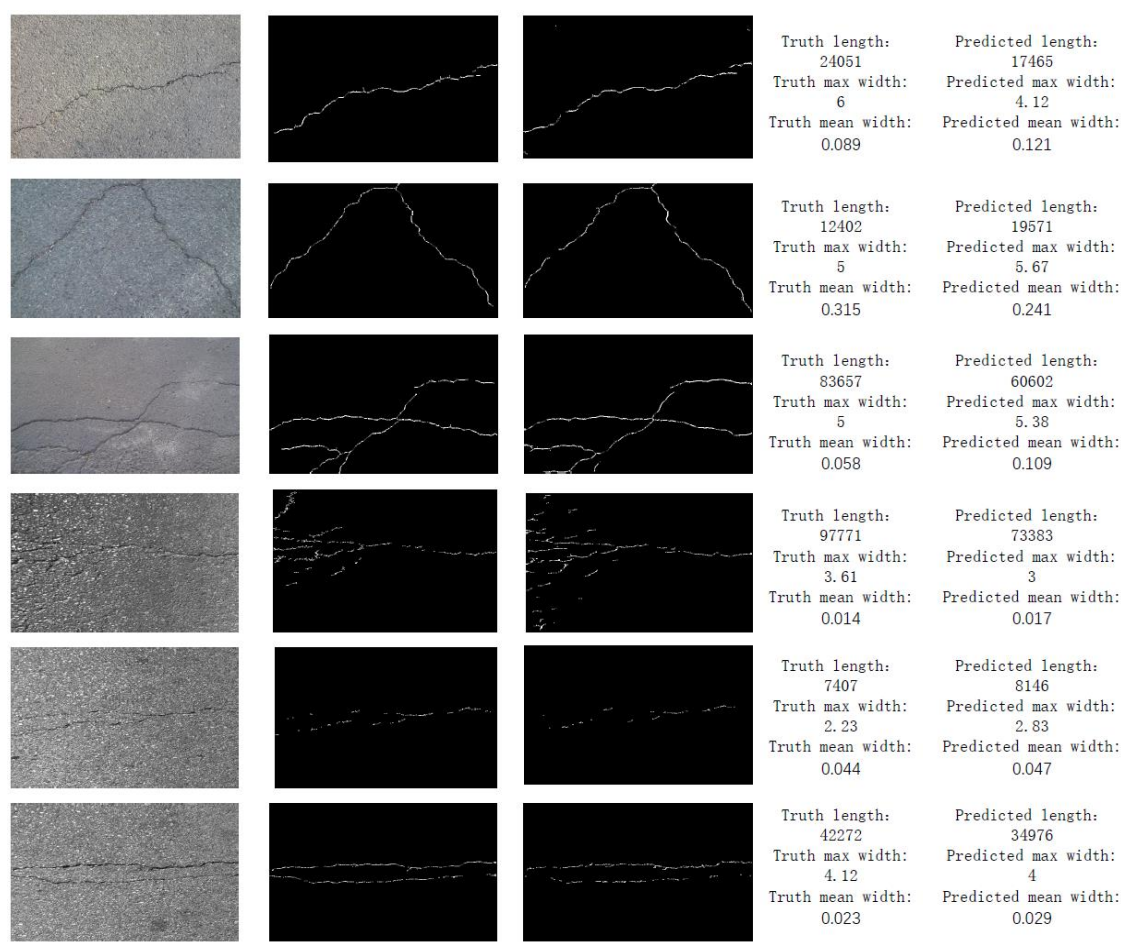

**Figure 8.** The experimental results show crack segmentation and crack skeleton based on public databases CFD and AigleRN. The numbers are in pixels.

## 4. Conclusions

The survey and analysis of road pavement distresses is an important issue for every Pavement Management System. All over the world, there have been many methods to gather information about the surface condition: some of them are only visual; others are based on advanced technologies. The first ones are economic but time-consuming and can be affected by errors due the operator subjectivity; the others are more reliable, even if the cost could be higher. In this paper, we propose an advanced method to evaluate the road pavement surface based an ensemble network of convolutional neural networks (CNN), based on probability fusion for automated pavement crack detection and measurement. The individual CNN designed improves the accuracy of crack detection by discarding pooling layers. The crack width and length can be measured based on predicted crack maps.

The experimental results were compared with existing databases and we found that precision, recall, and F1 had scores of 0.9552, 0.9521, and 0.9533 based on the CFD database, while the scores were 0.9302, 0.9166, and 0.9238 based on the AigleRN database. These results show that the proposed method outperforms the other methods. The algorithm adequately performs crack measurement: the length and the width of different crack types (complex, common, thin, and intersecting cracks) can be measured with satisfactory accuracy.

However, the proposed method is not able to perform end-to-end crack detection, and can only be employed to detect static images. Hence, we will explore the following in future work:

- We will explore end-to-end deep learning to create an automatic crack detection system.
- To date, many algorithms have detected cracks based on individual images. Therefore, we will explore the detection of cracks in video streaming.

**Author Contributions:** conceptualization, Z.F.; methodology, C.L.; software, Y.C.; validation, X.C. and G.Z.; formal analysis, C.L.; investigation, Y.C.; resources, C.L.; data curation, C.L.; writing—original draft preparation, C.L.; writing—review and editing, Z.F., P.D.M. and G.L.; visualization, C.L.; supervision, Z.F., P.D.M. and G.L.; project administration, Z.F., G.L.

**Funding:** This work was supported by the Science and Technology Planning Project of Guangdong Province of China under grant 180917144960530, by the Project of Educational Commission of Guangdong Province of China under grant 2017KZDXM032, by the State Key Lab of Digital Manufacturing Equipment and Technology under grant DMETKF2019020, and by the Project of Robot Automatic Design Platform combining Multi-Objective1 Evolutionary Computation and Deep Neural Network under grant 2019A 050519008.

**Acknowledgments:** We thank Giulia Del Serrone (Sapienza, University of Rome) for providing an accurate revision of English language and style.